\documentclass{article} % For LaTeX2e
\usepackage{iclr2027_conference,times}

\usepackage{amsmath,amsfonts,bm}

\def\eqref#1{equation~\ref{#1}}
\def\1{\bm{1}}

\DeclareMathAlphabet{\mathsfit}{\encodingdefault}{\sfdefault}{m}{sl}
\SetMathAlphabet{\mathsfit}{bold}{\encodingdefault}{\sfdefault}{bx}{n}

\newcommand{\E}{\mathbb{E}}

\usepackage{graphicx}
\usepackage{subcaption}
\usepackage{booktabs}
\usepackage{amsmath}
\usepackage{multirow}
\usepackage{tabularx}
\usepackage{longtable}
\usepackage{flafter}
\usepackage{placeins}
\usepackage{float}
\usepackage{url}
\usepackage{seqsplit}
\usepackage{tikz}
\usetikzlibrary{arrows.meta,positioning}
\usepackage[most]{tcolorbox}
\usepackage{hyperref}
\hypersetup{hidelinks,
  pdftitle={Does the Model Use the Feature? Separating Steering from Mechanism in LLMs},
  pdfauthor={Tong Che and Yilong Li}}
\newtcblisting{rubricprompt}[1]{
  enhanced,
  breakable,
  listing only,
  listing engine=listings,
  listing options={basicstyle=\ttfamily\fontsize{8.5}{10}\selectfont,breaklines=true,breakatwhitespace=true,breakautoindent=false,breakindent=0pt,columns=fullflexible,keepspaces=true,showstringspaces=false},
  colback=blue!3,
  colframe=blue!55!black,
  colbacktitle=blue!12,
  coltitle=blue!55!black,
  fonttitle=\sffamily\bfseries\small,
  title={#1},
  boxrule=0.55pt,
  arc=1.5mm,
  left=1.8mm,
  right=1.8mm,
  top=1.5mm,
  bottom=1.5mm,
  before skip=6pt,
  after skip=8pt
}

\newcommand{\Zfeat}{Z}

\newcommand{\pp}{\,\mathrm{pp}}

\title{Does the Model Use the Feature?\\
Separating Steering from Mechanism in LLMs}

\iclrfinalcopy

\author{Tong Che\thanks{Both authors contributed equally. Author order was determined by a roll of the dice.}\\
  {\normalfont NVIDIA Research}\\
  {\normalfont\ttfamily tongc@nvidia.com}
  \And
  Yilong Li\footnotemark[1]\\
  {\normalfont University of Wisconsin--Madison}\\
  {\normalfont\ttfamily yli758@wisc.edu}}
\date{}

\begin{document}

\raggedbottom

\maketitle
\pagestyle{plain}

\begin{abstract}
Internal features in LLMs are often interpreted as mechanisms when they track a concept and their manipulation changes a related behavior. Yet steering can push a feature far outside its natural range, where its effects need not reflect the model's own computation. We examine this inference and propose an empirical contract whose tests evaluate features at values observed on natural inputs. One test copies a feature's value from an input that shows a behavior into a matched input that does not (installation) or the reverse (removal); the other restores the feature after an upstream edit (downstream rescue). Installation measures how far the feature suffices for the behavior; removal and downstream rescue measure how much the model uses it. Applied to three kinds of representations, the two strengths separate sharply. The published unknown-entity latent strongly steers knowledge abstention, yet installing observed values from either published latent into matched prompts transfers only a small fraction of the natural known--unknown abstention contrast. Dense known--unknown directions show opposite asymmetries between installation and removal in Gemma and Llama, and how fully a released subject--verb agreement feature set reproduces and restores the behavior depends on how its values are written into the model. Tracking a concept and steering a behavior therefore do not by themselves show that the model uses a feature, and each conclusion holds only for the intervention tested.

\end{abstract}

\section{Introduction}
\label{sec:intro}

Internal features offer candidate explanations for how LLMs produce behavior. Prior work links refusal to a single direction~\citep{arditi2024refusal}, organizes sparse features into interpretable causal graphs~\citep{marks2024sparse}, and identifies sparse autoencoder (SAE) latents whose steering changes whether a model declines to answer questions about an entity~\citep{ferrando2025entity}, a behavior we call \emph{knowledge abstention}. These results invite a mechanistic reading: if a feature tracks a concept and editing it changes the associated behavior, the model may use that feature to produce the behavior. But control is not the same as mechanism. Steering can drive a feature far beyond its \emph{natural range}, the values it takes on natural inputs, and sufficiently large edits may recruit downstream pathways that the model does not ordinarily use. We therefore ask: \emph{when does feature-level control support a mechanistic interpretation of natural computation?}

\begin{figure}[t]
\centering
\begin{tikzpicture}[
  font=\scriptsize,
  inp/.style={draw, rounded corners=1.5pt, minimum width=1.7cm, minimum height=0.5cm, inner sep=2pt},
  model/.style={draw, fill=black!6, rounded corners=1.5pt, minimum width=2.3cm, minimum height=0.5cm},
  feat/.style={draw, circle, fill=white, minimum size=0.46cm, inner sep=0pt},
  set/.style={feat, draw=blue!60!black, very thick},
  edit/.style={draw, fill=black!30, minimum width=0.28cm, minimum height=0.5cm, inner sep=0pt},
  flow/.style={-{Stealth[length=1.4mm]}, semithick},
  copy/.style={-{Stealth[length=1.6mm]}, thick, blue!60!black},
]
\begin{scope}
  \node[anchor=west, font=\footnotesize] at (-0.05,1.6) {\textbf{a}\enspace Interchange (installation)};
  \node[inp] (a1) at (0.85,1.0) {unknown entity};
  \node[model] (am1) at (3.2,1.0) {};
  \node[feat] (az1) at (3.2,1.0) {$Z_1$};
  \node[anchor=west] (ao1) at (4.65,1.0) {abstains};
  \draw[flow] (a1) -- (am1);
  \draw[flow] (am1) -- (ao1);
  \node[inp] (a0) at (0.85,0) {known entity};
  \node[model] (am0) at (3.2,0) {};
  \node[set] (az0) at (3.2,0) {$Z_1$};
  \node[anchor=west] (ao0) at (4.65,0) {abstains?};
  \draw[flow] (a0) -- (am0);
  \draw[flow] (am0) -- (ao0);
  \draw[copy] (az1) -- node[right] {copy} (az0);
\end{scope}
\begin{scope}[xshift=7.0cm]
  \node[anchor=west, font=\footnotesize] at (-0.05,1.6) {\textbf{b}\enspace Downstream rescue};
  \node[inp] (b1) at (0.85,1.0) {unknown entity};
  \node[model] (bm1) at (3.2,1.0) {};
  \node[feat] (bz1) at (3.65,1.0) {$Z_1$};
  \node[anchor=west] (bo1) at (4.65,1.0) {abstains};
  \draw[flow] (b1) -- (bm1);
  \draw[flow] (bm1) -- (bo1);
  \node[inp] (b0) at (0.85,0) {unknown entity};
  \node[model] (bm0) at (3.2,0) {};
  \node[edit] (be) at (2.55,0) {};
  \node[below=1pt of be, inner sep=1pt] {upstream edit};
  \node[set] (bz0) at (3.65,0) {$Z_1$};
  \node[anchor=west] (bo0) at (4.65,0) {abstention returns?};
  \draw[flow] (b0) -- (bm0);
  \draw[flow] (bm0) -- (bo0);
  \draw[copy] (bz1) -- node[right] {restore} (bz0);
\end{scope}
\end{tikzpicture}
\caption{The two tests of the empirical contract, shown for entity familiarity and knowledge abstention. Each
row is a forward pass; circles mark the feature's value at its site. \emph{a:} Installation copies
the feature's natural value $Z_1$ from the input that shows the behavior (the target) into a
matched input that does not (the comparison) and asks whether behavior moves to the target level;
removal copies $Z_0$ the other way. \emph{b:}
Downstream rescue applies an upstream edit that changes the feature and the behavior, restores the
feature's natural value downstream while the edit stays active, and measures how much of the
behavioral change is reversed.}
\label{fig:tests}
\end{figure}
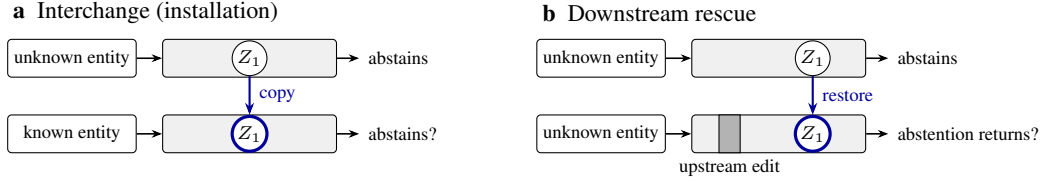

We study this question with an \emph{empirical contract} that distinguishes association, control, sufficiency, and use (Table~\ref{tab:evidence}). Association and control capture the familiar observations that a feature tracks a concept and that editing it changes behavior. The stronger claims ask whether the feature's natural values reproduce the behavior (\emph{sufficiency}) and whether the model's own computation depends on the feature (\emph{use}). We test both at values observed on natural inputs (Figure~\ref{fig:tests}). In an \emph{interchange} test~\citep{geiger2021causal}, paired inputs from two conditions exchange their feature values: \emph{installation} transfers the value associated with the target behavior, while \emph{removal} performs the reverse. In \emph{downstream rescue}, we edit the model upstream, restore the feature downstream, and measure how much behavior returns. Installation probes sufficiency; removal and rescue probe use.

\begin{table}[t]
\centering
\captionsetup{skip=4pt}
\caption{Four feature claims and the evidence used to assess them.}
\label{tab:evidence}
\footnotesize
\setlength{\tabcolsep}{5pt}
\renewcommand{\arraystretch}{1.05}
\begin{tabularx}{\textwidth}{@{}>{\raggedright\arraybackslash}p{1.7cm}>{\raggedright\arraybackslash}p{2.8cm}>{\raggedright\arraybackslash}X@{}}
\toprule
\textbf{Claim} & \textbf{Usual evidence} & \textbf{Contract evidence}\\
\midrule
Association & Development-set separation & Replication on held-out inputs with the site and
labels fixed.\\
Control & Steering at one strength & Predicted effect at a verified achieved value, compared with
the natural range.\\
Sufficiency & Nonzero installation effect & Sufficiency strength: the fraction of the natural
contrast reproduced by installing matched natural values.\\
Use & Nonzero removal effect & Use strength: the larger of the fraction of the contrast eliminated
by removal and the fraction of an upstream effect reversed by downstream rescue.\\
\bottomrule
\end{tabularx}
\end{table}

Our main audit focuses on the published entity-recognition latents of \citet{ferrando2025entity}. We reproduce their separation of known and unknown entities and the strong steering effect of the unknown-entity latent. Yet setting the latent to a value within its natural range barely changes knowledge abstention; the behavior increases only when the latent is pushed beyond that range (Figure~\ref{fig:dose}). Likewise, installing the known-entity latent's common-name value into a prompt that names the same entity by a rare alias does not recover the accuracy gain of the common name. An independently fitted refusal direction serves as a positive control, showing that our tests detect effects at natural values when they are present. To separate a failure of the SAE basis from a failure of the underlying concept, we also test dense known--unknown directions defined by class-mean differences. Their behavior is asymmetric across models: Llama shows strong installation but weak removal and rescue, whereas Gemma shows stronger removal but weak installation and rescue. By contrast, a released subject--verb agreement feature set spanning multiple layers shows high sufficiency and use strengths on held-out vocabulary when its values are written through the SAE decoders (the \emph{decoder update}).

The same feature set shows why the update rule, the procedure used to write a value into the
model, is part of the intervention claim. Both the decoder update and the encoder-constrained update,
which matches the requested feature values more closely, recover the preferred verb on
almost all pairs. Yet the decoder update yields a much higher sufficiency strength and stronger rescue: downstream
layers read the residual stream, not the encoder's output, and the two updates change the
residual stream differently. The entity results make
the same point about the representation: an SAE latent and a dense direction with the same
semantic label behave differently under the same tests. The contract therefore specifies the
representation, site, edited tokens, update rule and behavioral measure together.

Our contributions are:
\begin{itemize}
\item \textbf{An empirical contract for feature claims} that measures the strength of sufficiency
and use at values observed on natural inputs (Section~\ref{sec:method}).
\item \textbf{An audit of entity-familiarity representations}: the published SAE latents steer
knowledge abstention, but paired interchange transfers only a small fraction of the natural
known--unknown abstention contrast; dense directions dissociate installation from removal
(Sections~\ref{sec:eval-assoc}--\ref{sec:eval-interchange}).
\item \textbf{A positive case that depends on the update rule}: the agreement feature set of
\citet{marks2024sparse} reaches its highest sufficiency and rescue only under the decoder update
(Sections~\ref{sec:eval-interchange} and~\ref{sec:eval-operator}).
\end{itemize}

\section{Related work}
\label{sec:related}

\noindent\textbf{SAE features and semantic interpretations.}
Sparse autoencoders decompose transformer activations into sparse learned features, many with interpretable semantic descriptions
\citep{cunningham2023sparse,bricken2023monosemanticity,templeton2024scaling}.
Resources such as Gemma Scope provide pretrained JumpReLU dictionaries for feature-level analysis
\citep{lieberum2024gemmascope,rajamanoharan2024jumprelu}.
\citet{ferrando2025entity} identify entity-recognition latents that distinguish known from unknown entities and connect them to knowledge abstention and factual recall through steering, weight orthogonalization, and activation patching. Other work probes feature interpretations through behavioral falsification \citep{ma2026reasoning}, and \citet{paulo2025different} find that feature identity can vary across retraining seeds, especially for TopK dictionaries. Entity-feature audits further separate detection from control: fabricated-entity detection and refusal steering can occupy distinct directions \citep{galeone2026detection}, while a Qwen replication finds no calibration gain from single-feature steering and increased hallucination after multi-feature ablation \citep{vicentino2026hallucination}. \citet{cui2026unreliable} recover suppressed behavior through residual-space edits while holding targeted SAE features fixed. \citet{makelov2025principled} benchmark SAE disentanglement and control on indirect-object identification against supervised task-specific dictionaries. Our audit is similarly task-grounded, but tests fixed candidate features through installation, removal and rescue at values observed on natural inputs.

\noindent\textbf{Steering, activation patching, and intervention validity.}
Steering vectors can control behavior across forced-choice tasks and open-ended traits
\citep{rimsky2024steering,chen2025persona}, though SAE steering may underperform simple prompting
\citep{wu2025axbench}. For refusal, \citet{arditi2024refusal} derive a harmful--harmless direction and control refusal through addition and directional ablation. Later work identifies refusal concept cones and geometrically distinct directions with similar refusal--over-refusal trade-offs
\citep{wollschlager2025geometry,joad2026refusal}. We distinguish a representation from the edit used to intervene on it: addition, directional ablation, and replacement with an observed value are different interventions. Interchange interventions replace internal states with values from another input to test causal abstractions and trace computations
\citep{geiger2021causal,geiger2023abstraction,vig2020causal,meng2022locating}, and have been used for attribute disentanglement \citep{huang2024ravel}, interpretation validity \citep{li2026bucketing}, and steerable causal subspaces \citep{bao2026faithful}. We apply interchange to fixed candidate representations and compare the installed behavior with that of the matched natural input. Our downstream-rescue test follows denoising activation patching \citep{heimersheim2024patching}, restoring the candidate state while keeping the upstream edit active. Patching conclusions can depend on the metric, corruption scheme, and ablation rule
\citep{zhang2024bestpractices,miller2024notrobust}; causal localization also need not identify the best sites for knowledge editing \citep{hase2023localization}, and subspace patching can activate dormant pathways rather than restore the model's usual computation \citep{makelov2023illusion}. Finally, feature interventions can exceed typical activation levels
\citep{templeton2024scaling,ameisen2025circuit} or reach residual states not produced by any prompt
\citep{mishra2026nonsurjective}. We therefore measure how far steering moves feature values beyond their observed natural range.

\noindent\textbf{Behavioral dependence and causal claims.}
Amnesic probing tests behavioral dependence by removing decodable information and measuring the resulting behavioral change \citep{elazar2021amnesic}. Work on grammatical number uses usage probes
\citep{lasri2022usage}, causal mediation \citep{finlayson2021agreement}, and counterfactual interventions
\citep{ravfogel2021counterfactual} to test which representations contribute to subject--verb agreement; we use the released agreement annotations of \citet{marks2024sparse} as a candidate positive case. \citet{joshi2026causality} organize mechanistic interpretations within a causal hierarchy, and \citet{lin2026identification} emphasize the assumptions required for causal validation. Our empirical contract separates association, control, sufficiency and use, and tests the last two only at values observed on natural inputs.

\section{Background: the audited features}
\label{sec:background}

\noindent \textbf{Entity-recognition latents.}
\citet{ferrando2025entity} construct a Wikidata dataset of basketball players, movies, cities,
and songs and prompt the Gemma~2~2B base model to complete their attributes, e.g., ``The movie
12 Angry Men was directed by''. An entity is \emph{known} if the model answers at least two
attributes correctly and \emph{unknown} if it answers all incorrectly; intermediate cases are
discarded. At the entity's last token, which we call the \emph{entity token}, they select latents
from the Gemma Scope dictionary for Gemma~2~2B whose firing rates best separate known from unknown
entities across all four types. As in their steering experiments, we audit the resulting
known-entity latent L13/7957 and unknown-entity latent L15/11898 in Gemma~2~2B-IT, using the
released entities and labels.

\noindent \textbf{Knowledge abstention and steering.}
For behavioral evaluation, these completions become questions such as ``Who directed the
movie 12 Angry Men?''. We use \emph{knowledge abstention} for responses that give no substantive
answer because they express uncertainty, claim inability, decline, or reject the entity or premise.
It differs from refusing harmful requests and from instruction refusal, which our positive control
measures (Section~\ref{sec:eval}).
\citet{ferrando2025entity} identify such responses by string matching and steer by adding the
latent decoder direction $d_j$ to the residual stream,
$r \leftarrow r+\alpha d_j$, at the entity token and subsequent end-of-instruction tokens, with
$\alpha$ roughly twice the residual-stream norm. Steering with the unknown-entity latent causes
near-universal abstention on unknown-entity questions, while the known-entity latent induces
hallucination about a non-existent player the model would otherwise decline to answer.

\noindent \textbf{Subject--verb agreement features.}
\citet{marks2024sparse} annotate SAE features in Pythia-70M-deduped that mediate subject--verb
agreement, where the model must prefer the verb form matching the subject's grammatical number,
e.g., ``is'' after the prefix ``The nurse'' and ``are'' after ``The nurses''.
We treat $147$ annotated features across six residual-stream sites as one feature set. Because the
released test split reuses the vocabulary used for annotation, we evaluate on newly constructed
nouns (Appendix~\ref{app:interchange}). Table~\ref{tab:representations} summarizes all audited
representations.

\begin{table}[t]
\centering
\captionsetup{skip=4pt}
\caption{Representations audited in this work; Section~\ref{sec:eval} adds a refusal direction as
a positive control.}
\label{tab:representations}
\footnotesize
\setlength{\tabcolsep}{4pt}
\renewcommand{\arraystretch}{1.1}
\begin{tabularx}{\textwidth}{@{}>{\raggedright\arraybackslash}p{3.9cm}>{\raggedright\arraybackslash}p{2.9cm}>{\raggedright\arraybackslash}p{2.9cm}>{\raggedright\arraybackslash}X@{}}
\toprule
\textbf{Representation} & \textbf{Model} & \textbf{Site} & \textbf{Origin}\\
\midrule
Known- and unknown-entity latents (L13/7957, L15/11898) &
Gemma~2~2B-IT &
entity token, layers 13 and 15 &
\citet{ferrando2025entity}\\

Known--unknown direction &
Gemma~2~2B-IT &
entity token, layer 15 &
class-mean difference fit on disjoint entities\\

Known--unknown direction &
Llama~3.1~8B-Instruct &
entity token, layer 15 &
class-mean difference fit on disjoint entities\\

Agreement feature set ($147$ features) &
Pythia-70M-deduped &
final prefix token, six sites &
annotations of \citet{marks2024sparse}\\
\bottomrule
\end{tabularx}
\vspace{=0.5ex}
\end{table}

\vspace{-0.5ex}
\section{The empirical contract and its tests}
\label{sec:method}
\vspace{-0.5ex}
Because sufficiency and use concern the model's own computation, the contract tests them only by
setting the feature to values observed on natural inputs, and asks whether these edits reproduce
the natural difference between conditions, not merely whether behavior is sensitive to them. A
candidate feature state $\Zfeat(x)$ is the value of a single latent or direction at a specified
site, or the joint state of a feature set across several sites; $B$ is the behavior it is meant to
explain. The four claims of the contract (Table~\ref{tab:evidence}) ask progressively more of the
feature: the usual evidence for association and control is that the feature separates the two
conditions and that steering it changes $B$, and the contract adds the requirements in the last
column of Table~\ref{tab:evidence}. Each claim is measured by its own test, and a weak result
limits the evidence for that claim without ruling out other causal roles of the feature.

\vspace{-0.5ex}
\subsection{What we estimate}
\label{sec:estimands}
\vspace{-0.5ex}
Each item $e$ pairs a comparison input $c_0$ with a target input $c_1$ on which $B$ is expected to
be higher. In the entity task, $c_1$ asks about an entity labeled unknown and $c_0$ about a
matched entity labeled known, using the same prompt type; $B$ is knowledge abstention. In the
agreement task, $c_0$
and $c_1$ differ only in the subject's grammatical number (plural in $c_0$, singular in $c_1$), and
$B=\ell(v_{\mathrm{sg}})-\ell(v_{\mathrm{pl}})$ is the difference between the logits $\ell$ of the
singular and plural verb forms. Let $Z_j=\Zfeat(c_j)$ be the feature's natural value on each
input, and let $Y(c,z)$ be the behavior on input $c$ when the update rule sets the feature to $z$;
for a multi-site feature set, $z$ is the tuple of requested site values, and the update rule
writes each component at its site as the forward pass reaches it. The natural contrast is
$\Delta_{\mathrm{nat}}=\E[Y(c_1,Z_1)-Y(c_0,Z_0)]$, with intervals clustered by paired entity or noun
group.

\textbf{Interchange.}\; Installation runs the comparison input with the target's value,
$Y(c_0,Z_1)$, and removal runs the target input with the comparison's value, $Y(c_1,Z_0)$. We
express both as fractions of the natural contrast:
\[
\rho_{\mathrm{install}}=\frac{\E[Y(c_0,Z_1)-Y(c_0,Z_0)]}{\Delta_{\mathrm{nat}}},\qquad
\rho_{\mathrm{remove}}=\frac{\E[Y(c_1,Z_1)-Y(c_1,Z_0)]}{\Delta_{\mathrm{nat}}}.
\]
We take $\rho_{\mathrm{install}}$ as the \emph{sufficiency strength}: a value of $1$ means that
installing the target's natural value reproduces the target's mean behavior, and the remaining
installation gap, $\E[Y(c_1,Z_1)-Y(c_0,Z_1)]=(1-\rho_{\mathrm{install}})\Delta_{\mathrm{nat}}$,
is the part of the
contrast the feature does not carry. As sensitivity analyses, installation and removal can also
set the feature to an observed class mean instead of the matched value.

\textbf{Downstream rescue.}\; Rescue applies an upstream edit that changes both $Z$ and $B$ on the
target input, then restores the feature to its natural value downstream while the edit remains
active. With $Y_U$ the outcome under the upstream edit alone and $Y_{U+R}$ the outcome after
restoration, $\rho_{\mathrm{rescue}}=-\E[Y_{U+R}-Y_U]/\E[Y_U-Y(c_1,Z_1)]$ is the fraction of the
upstream effect that restoration reverses. Removal asks whether behavior depends on the feature's
natural value, and rescue asks whether an upstream effect on behavior passes through the feature.
Each is sufficient evidence of use but neither is necessary: when several components carry the same
information, removing one leaves behavior intact while restoring one recovers it, and when several
components are jointly required, the reverse holds \citep{heimersheim2024patching}. We therefore
take the larger of $\rho_{\mathrm{remove}}$ and $\rho_{\mathrm{rescue}}$ as the \emph{use strength},
and report every ratio with its $95\%$ interval rather than thresholding it.

\textbf{Update rules.}\; Both strengths are conditional on the update rule. For SAE features $S$ with
encoder output $a=\operatorname{Enc}(r)$ and decoder columns $W_{\mathrm{dec},S}$, the
\emph{decoder update} writes the requested values $a^\star_S$ along the decoder columns, and for a
dense direction $v$ with requested projection $q^\star$ we set the projection directly:
\[
r'=r+W_{\mathrm{dec},S}\,(a^\star_S-a_S),\qquad
r'=r+\frac{(q^\star-v^\top r)\,v}{\lVert v\rVert_2^2}.
\]
The paired writes for individual Gemma latents use the same gain-corrected setter as the dose study
(Appendix~\ref{app:dose-confirmation}); the feature-set decoder update above is unchanged. For
L15/11898, the residual write along $d_j$ scales the requested preactivation change by
$1/\hat g_j\approx3.5$. The requested value is observed on a donor input in the encoder readout;
the edited residual need not be a donor-observed state.
Because re-encoding $r'$ need not return $a^\star_S$, the \emph{encoder-constrained update}
instead solves for a residual change $\delta$ whose re-encoded values match the request,
$\max_{k\in S}|\operatorname{Enc}_k(r+\delta)-a^\star_k|\le 10^{-3}$; we report the two
feature-set updates separately. For a dense direction, we set the projection directly rather than
write through SAE decoders, which would change large residual components orthogonal to the
direction (Appendix~\ref{app:interchange-operators}). We run no-op controls and report achieved
values after re-encoding; we report no strength for
an edit whose achieved value falls outside a preset tolerance of the request.

\textbf{Scope.}\; The subspace-patching illusion of \citet{makelov2023illusion} arises within MLP
hidden activations, where it exploits the nullspace of the MLP output projection. Our edits target
residual-stream representations, a location they argue is less prone to such illusions, so their
construction is not a direct counterexample to these tests. Natural feature values still need not
produce naturally realizable joint residual states, so a high strength is evidence under the specified
representation, sites, edited tokens and update rule, not a general guarantee of mechanistic
identification.

\section{Applying the empirical contract to published features}
\label{sec:eval}

\paragraph{Setup.} We intervene on Gemma~2~2B-IT, Llama~3.1~8B-Instruct and
Pythia-70M-deduped, generating open-ended responses from Gemma and Llama greedily. Features, sites,
update rules, targets and measures are fixed on development data before confirmation
runs on held-out data (Appendices~\ref{app:registration} and~\ref{app:interchange}), which also
give the datasets, hardware, software and intervention settings. We run the tests of
Section~\ref{sec:estimands} only when the interval for $\Delta_{\mathrm{nat}}$ excludes zero.

\paragraph{Labels.} Responses to entity questions are labeled by OpenAI \texttt{gpt-5.6-terra}
under a fixed six-category rubric: explicit refusal, invalid entity or premise, knowledge
uncertainty and claimed inability count as knowledge abstention, whereas substantive answers and
other responses do not. On $574$ responses annotated independently by two humans, these labels
reach a balanced accuracy of $0.94$ $[0.92,0.96]$ for knowledge abstention
(Appendix~\ref{app:primary-endpoint}). A Qwen2.5-3B-Instruct judge with fixed prompts (the
\emph{Qwen judge}) scores instruction refusal in the refusal control and a few sensitivity
analyses.

Section~\ref{sec:eval-assoc} reproduces the usual evidence for association and control, and
Section~\ref{sec:eval-dose} tests the latents at values within and far beyond their natural range.
Section~\ref{sec:eval-interchange} applies interchange and downstream rescue, and
Section~\ref{sec:eval-operator} varies the update rule and the edited tokens.

\subsection{The published latents retain association and strong steering}
\label{sec:eval-assoc}
We begin with the evidence on which the published interpretation rests. On the publication's
held-out test split of $800$ entities, the known-entity latent L13/7957 and the unknown-entity
latent L15/11898 separate known from unknown entities with macro areas under the ROC curve
(AUROCs) of $0.874$ and $0.745$. We then apply the published steering to the unknown-entity latent at $\alpha=400$, the
lower end of the reported $400$--$550$ range, from the entity token to the end of the prompt. On
$400$ unknown-entity questions, steering raises knowledge abstention from $33.0\%$ to $79.0\%$, a
paired change of $+46.0\pp$ $[40.5,51.5]$, with an increase for every entity type
(Appendix~\ref{app:primary-endpoint}).

\subsection{The latents barely change behavior within their natural range}
\label{sec:eval-dose}
Steering shows that behavior is sensitive to an edit, not that the model relies on the values the
feature takes naturally. We therefore set the unknown-entity latent at the entity token only to one
of two targets: a natural target near the $95$th percentile of its unedited activations, and a high
target at the value the published steering reaches, beyond every unedited activation. On $400$ new
unknown-entity questions, we compare knowledge abstention under each target with the unedited
baseline; the achieved activations match the requests (Appendix~\ref{app:dose-confirmation}).

\begin{figure}[t]
\centering
\captionsetup{skip=4pt}
\includegraphics[width=0.94\textwidth]{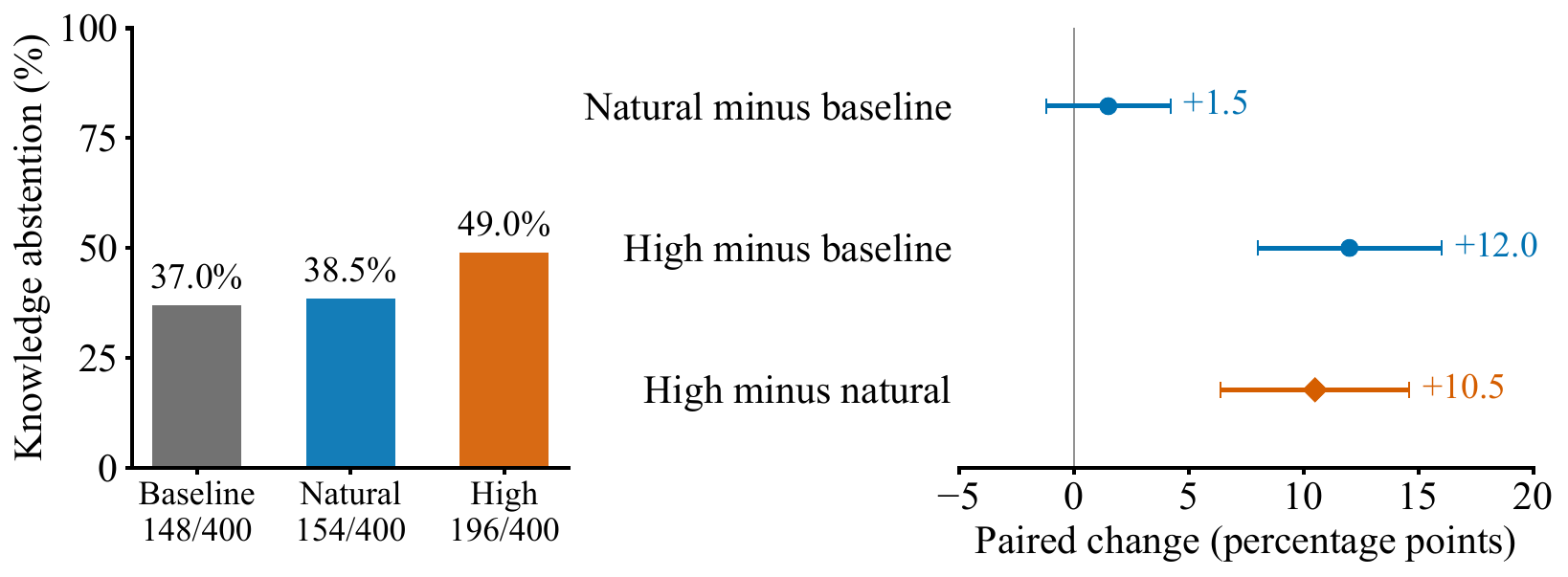}
\caption{Only the high target, beyond every unedited activation, clearly raises knowledge
abstention. The unknown-entity latent is left unedited (baseline) or set to the natural or the
high target on $400$ new unknown-entity questions. Left: abstention counts and rates. Right: paired
changes with $95\%$ intervals.}
\label{fig:dose}
\end{figure}

With the feature, site, token, update rule and prompts held fixed, only moving the latent beyond
its natural range clearly changes behavior (Figure~\ref{fig:dose}). The natural target changes
knowledge abstention by only $+1.5\pp$ $[-1.2,4.2]$, whereas the high target raises it by
$+12.0\pp$ and exceeds the natural target by $+10.5\pp$ $[6.4,14.6]$.

A natural input change exposes the same failure in the known-entity latent. The same question can name an entity in more than
one way, as in ``What genre is \emph{Battlefield Play4Free}?'' versus ``What genre is
\emph{BFP4F}?'' On $500$ held-out alias pairs, asking with the entity's common name rather than a
rare alias raises factual accuracy by $11.2\pp$ $[7.6,14.8]$, and common names also raise the
known-entity latent. If the latent carried the accuracy gain, installing its common-name value in
the rare-name prompt should recover the gain. Copying the full residual state at the same site
recovers the gain, changing accuracy by $10.6\pp$ $[7.2,14.0]$, but copying only the
latent's value changes it by $-0.2\pp$ $[-2.0,1.6]$ (Appendix~\ref{app:registration}).

As a positive control, a refusal direction fit following \citet{arditi2024refusal} shows that an
edit within the natural range can change behavior: setting its projection to the refusal-class mean
raises instruction refusal by $+29.0\pp$ $[18,39]$ on $100$ held-out prompts
(Appendix~\ref{app:registration}).

\vspace{-0.5ex}
\subsection{Installation, removal and rescue dissociate across representations}
\label{sec:eval-interchange}
\begin{figure}[!t]
\centering
\captionsetup{skip=4pt}
\includegraphics[width=0.84\textwidth]{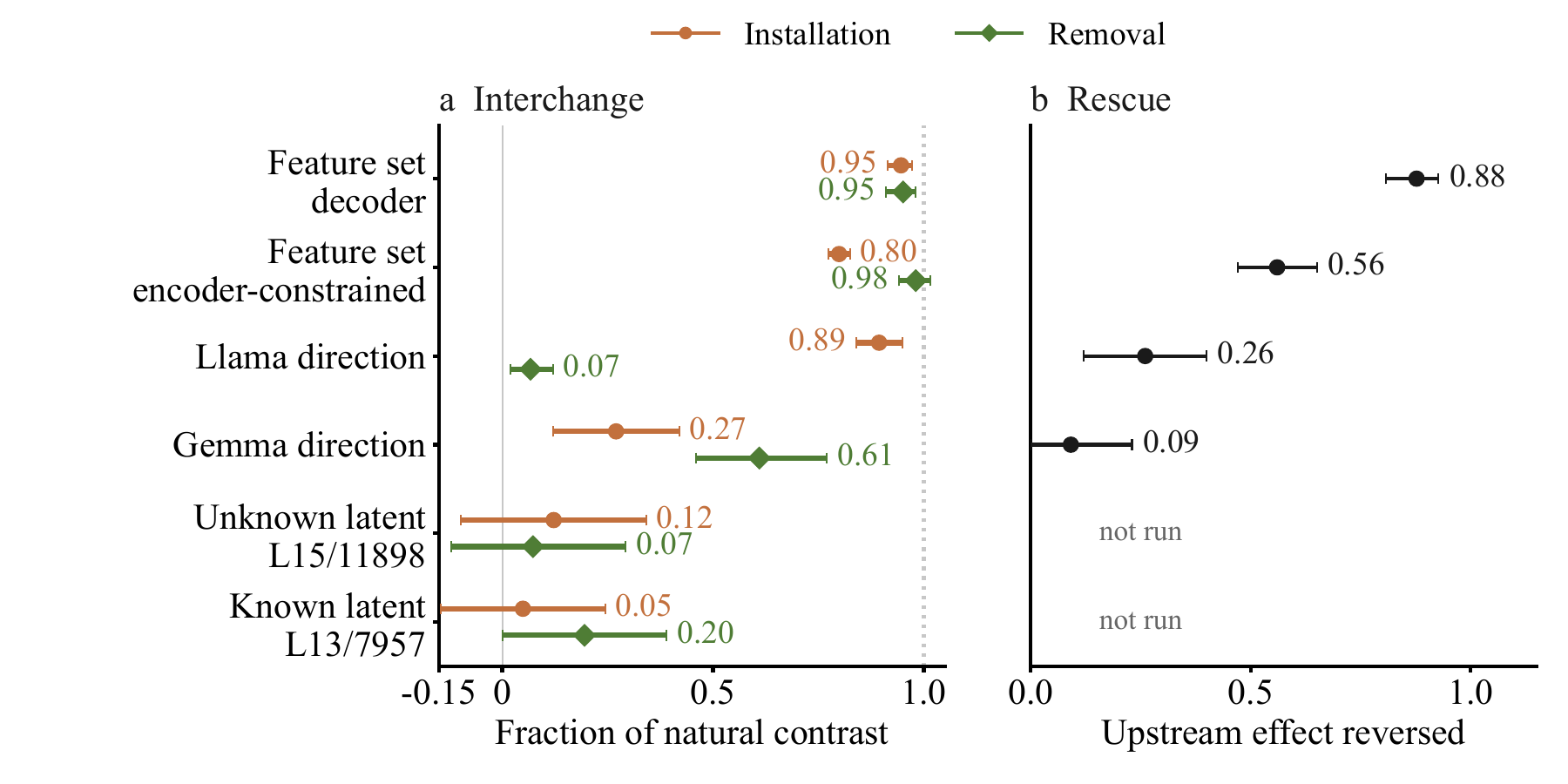}
\caption{Interchange and rescue across representations. \emph{a:} Fractions of the natural contrast
transferred by installation and eliminated by removal (95\% intervals; dotted line: full
contrast). \emph{b:} Fraction of the upstream effect reversed (95\% intervals). The two
feature-set rows use the same $58$ noun groups and upstream
edit, changing only the update rule. Both directions use the entity token. Rescue was not run for
the two Gemma latents.}
\label{fig:interchange}
\vspace{-0.5ex}
\end{figure}

\vspace{-0.5ex}
On matched known--unknown questions, paired interchange with the published Gemma latents transfers
$0.12$ of the natural knowledge-abstention contrast for L15/11898 and $0.05$ for L13/7957 under
installation; removal eliminates $0.07$ and $0.20$, respectively (Table~\ref{tab:interchange},
Figure~\ref{fig:interchange}). None of the four paired-effect intervals excludes zero.
Rescue was not run for either latent, so their use strength is known only to be at least their
removal fraction. To
show what high sufficiency and use strengths look like, we apply interchange and rescue to the
released agreement feature set on
held-out nouns; to test whether these small effects reflect the SAE basis rather than the
familiarity concept, we apply them to Gemma and Llama known--unknown directions
(Section~\ref{sec:estimands}). Both directions use the last entity token, matching the readout site
of \citet{ferrando2025entity} in Gemma and Llama.

\noindent\textbf{Pairs and upstream edits.}
For the agreement feature set, singular and plural prefixes from $58$ held-out noun groups form $232$ pairs. For the directions, known entities are matched to unknown entities of the same prompt type, yielding $400$ Llama and $148$ Gemma pairs, and interchange copies each example's observed projection to its match. The published Gemma latents use the same $148$ pairs. For rescue, the agreement upstream edit modifies $60$ annotated embedding features upstream of all six feature-set sites. For the directions, the upstream edit sets the projection of unknown prompts on an upstream direction to the known-class mean at the entity token (Gemma: layer 8; Llama: development-selected layer 12; Appendix~\ref{app:interchange}).

\begin{table}[htbp]
\centering
\captionsetup{skip=4pt}
\caption{Sufficiency and use strengths. Entries are fractions of each natural contrast; the
sufficiency strength is $\rho_{\mathrm{install}}$, and the use strength is the larger of
$\rho_{\mathrm{remove}}$ and $\rho_{\mathrm{rescue}}$. Figure~\ref{fig:interchange} shows $95\%$
intervals; Appendix~\ref{app:interchange} reports exact intervals and other diagnostics.}
\label{tab:interchange}
\footnotesize
\setlength{\tabcolsep}{4pt}
\renewcommand{\arraystretch}{1.08}
\begin{tabularx}{\textwidth}{@{}>{\raggedright\arraybackslash}Xcccc@{}}
\toprule
& Sufficiency & \multicolumn{3}{c}{Use} \\
\cmidrule(lr){2-2}\cmidrule(l){3-5}
Representation & strength $\rho_{\mathrm{install}}$ & removal $\rho_{\mathrm{remove}}$ & rescue $\rho_{\mathrm{rescue}}$ & strength \\
\midrule
Agreement feature set, decoder & $0.946$ & $0.951$ & $0.877$ & $0.951$ \\
Agreement feature set, encoder-constrained & $0.799$ & $0.981$ & $0.560$ & $0.981$ \\
Llama direction & $0.89$ & $0.07$ & $0.26$ & $0.26$ \\
Gemma direction & $0.27$ & $0.61$ & $0.09$ & $0.61$ \\
Gemma latent L15/11898 & $0.122$ & $0.073$ & --- & ${\ge}\,0.073$ \\
Gemma latent L13/7957 & $0.049$ & $0.195$ & --- & ${\ge}\,0.195$ \\
\bottomrule
\end{tabularx}
\par\vspace{3pt}
\begin{minipage}{\textwidth}
Natural contrasts: agreement feature set $7.32$ verb-logit units; Llama $+44.75\pp$ ($179/400$); Gemma
$+27.70\pp$ ($41/148$).
\end{minipage}
\end{table}

\noindent\textbf{Agreement feature set.} Written through its SAE decoders, the feature set nearly
reproduces both natural conditions (Table~\ref{tab:interchange}). Installation and removal each
account for about $0.95$ of the natural contrast, and downstream restoration reverses $0.88$ of the
upstream effect. Both strengths are far above those of the entity representations.

\noindent\textbf{Llama direction.} Installation transfers $0.89$ of the natural contrast in
knowledge abstention, a sufficiency strength close to that of the agreement feature set. Removal
eliminates only $0.07$, however, and downstream restoration reverses $0.26$ of the upstream effect,
so the use strength is low ($0.26$). The direction's observed value nearly suffices to produce the
target behavior, yet behavior barely depends on it at this site.

\noindent\textbf{Gemma direction.} Gemma shows the opposite asymmetry. At the entity token, removal
eliminates $0.61$ of the natural contrast, but installation transfers only $0.27$, and restoring
the direction downstream reverses only $0.09$ of the upstream effect, so its use strength ($0.61$)
exceeds its sufficiency strength. Knowledge abstention at this site is sensitive to removal, but the
direction's observed value is not enough to reproduce the target behavior, and the upstream edit
acts mostly through other routes.

\subsection{The intervention claim depends on the update rule and edited tokens}
\label{sec:eval-operator}
A feature value can be written into the model in more than one way, and the ways differ in what
else they change. For the agreement feature set, matching the requested feature values more
exactly does not make the behavior more faithful. After re-encoding, the encoder-constrained
update matches the requested values several orders of magnitude more closely than the decoder
update and recovers the target verb on about as many pairs
(Appendix~\ref{app:interchange-operators}), yet the decoder update yields a higher sufficiency
strength and stronger rescue. Removal, and with it the use strength, is nearly invariant to the
update rule ($0.951$ versus $0.981$), whereas
installation ($0.946$ versus $0.799$) and especially rescue ($0.877$ versus $0.560$) are not
(Table~\ref{tab:interchange}).

The refusal control shows that the edited tokens matter as well
(Table~\ref{tab:refusal-operators}). Zeroing the direction's projection at one site and token
lowers refusal far less than ablating the direction at every layer and position, and removal at
that site also lowers refusal.

\begin{table}[htbp]
\centering
\captionsetup{skip=4pt}
\caption{Removal at one site and token lowers refusal, whereas ablation at every layer and position
suppresses it much more. The same refusal direction and $100$ prompts throughout; instruction refusal is
scored by the Qwen judge, and brackets are $95\%$ intervals.}
\label{tab:refusal-operators}
\small
\begin{tabular}{@{}llr@{}}
\toprule
Edit & Scope & Change in refusal (pp)\\
\midrule
Removal (projection set to the answer-class mean) & One site and token & $-6.0$ $[-11,-2]$\\
Projection set to zero & One site and token & $-5.0$ $[-10,-1]$\\
Directional ablation & All layers and positions & $-45.0$ $[-55,-35]$\\
\bottomrule
\end{tabular}
\end{table}

\section{Discussion}
\label{sec:discussion}

\paragraph{Why is published steering stronger than paired interchange?} Steering and interchange
ask different questions. The published steering adds about
twice the residual-stream norm along the decoder direction and reaches activations far outside the
natural range (Section~\ref{sec:eval-dose}). An edit of that size can change knowledge abstention
through any downstream computation that is sensitive to the direction, including computations
that natural variation in the latent never engages. Interchange asks instead whether values
observed on natural inputs carry the behavioral difference; the published latents transfer little
of it, with modest removal for the known-entity latent. The dissociation parallels the gap between
localization and editing reported by
\citet{hase2023localization} and the dormant pathways of \citet{makelov2023illusion}: an
intervention can succeed without identifying the representation the model uses. Studies that
validate a feature interpretation by steering therefore test control, not sufficiency or use.

\paragraph{What do the asymmetries of the dense directions show?} Installation and removal can
dissociate either way, so sufficiency and use must be tested separately. The Llama direction
nearly suffices, yet removing it at the entity token barely lowers knowledge abstention; Gemma shows the
opposite asymmetry, and downstream rescue is weak in both. Previous ablation studies find that
later components can compensate for edited attention layers or heads
\citep{mcgrath2023hydra,rushing2024selfrepair}. Similar compensation could make Llama's $0.07$
removal effect and the weak rescues in both models understate the directions' contribution. The refusal direction likewise installs
refusal at one token, but removal there suppresses it weakly, whereas ablation at every layer and
position does \citep{arditi2024refusal}. A single-token edit thus probes one location of a
computation that may span tokens and layers.

\paragraph{Why does the update rule change the result?} The two feature-set updates request the
same values but change different residual-stream components. Downstream layers read the residual
stream, not the SAE encoder's output, so a write whose re-encoded values match the request more
closely need not reproduce the natural computation more closely: the encoder-constrained update is
the more accurate in feature space and the less accurate in behavior\footnote{This comparison
concerns installation and rescue. Removal is slightly larger for the encoder-constrained update
($0.981$ versus $0.951$), as is use strength under our definition.}
(Section~\ref{sec:eval-operator}).

\vspace{-0.6ex}
\section{Conclusion}
\label{sec:conclusion}
\vspace{-0.6ex}

Does the model use the feature? Association and control alone cannot answer this. We reproduce the
known--unknown separation and the strong steering effect of the published entity-recognition
latents. Under paired interchange, however, these single latents carry only a small fraction of the
natural known--unknown abstention contrast, with modest asymmetric removal for the known-entity latent.
Dense known--unknown directions show opposite asymmetries between installation and removal in the
two models, and the agreement feature set reaches its highest sufficiency and rescue only under the
decoder update. Feature claims should therefore report the values steering reaches relative to the
natural range, measure sufficiency by installing values observed on natural inputs, measure use
through removal and downstream rescue, and specify the representation, site, edited tokens, update rule
and behavioral measure together.

\subsubsection*{Reproducibility statement}
An accompanying artifact contains the input sets, raw generations, judge outputs, intervention
diagnostics, analysis code and figure-generation scripts needed to reproduce the reported results.
The complete experimental codebase will be released upon publication.

\subsubsection*{AI use statement}
We used Anthropic Claude and OpenAI GPT/Codex systems for literature discovery, code implementation and review, analysis support, and paper polishing. We verified cited sources and scientific claims, reviewed generated code and
reproduced the reported analyses and software tests.

\bibliography{references}
\bibliographystyle{plainnat}

\clearpage
\appendix
\section{Direction fitting, refusal control and labeling}
\label{app:required}

\paragraph{Direction fitting and sites.} \citet{ferrando2025entity} read the final token of each
entity; their top Llama~3.1~8B unknown-entity latent is at layer 15. Our known--unknown directions
are separately fitted class-mean differences on disjoint entities at that layer and token in both
models (Table~\ref{tab:representations}).

\paragraph{Labels.}
\label{app:primary-endpoint}
We use OpenAI \texttt{gpt-5.6-terra} only as a response classifier; it does not generate any task
responses. The adjudicator receives the question, reference answer and generated response, but not
the intervention condition, model condition, baseline-versus-edited status or expected label.
Reasoning is disabled, sampling is off, temperature is $0$ where supported by the API and the
maximum output length is $32$ tokens. A structured output schema restricts the response to one of
six labels. An invalid output is retried once with the same prompt and otherwise mapped to
\texttt{OTHER}. The adjudicator assigns one of six categories: substantive answer, explicit
refusal, invalid entity or premise, knowledge uncertainty, claimed inability, or other. On entity
questions, explicit refusal, invalid entity or premise, knowledge uncertainty and claimed
inability count as knowledge abstention when the response supplies no substantive answer. Safety
and instruction refusal use separate binary mappings. Identical
question--reference--response triples are adjudicated once and share that label.

\paragraph{Rubric development and human comparison.} The rubric was developed on the
$88$-question dose development set and its responses, together with a separate set of $574$
responses annotated by two humans blind to intervention condition. Against the human labels,
six-class macro-F1 is $0.86$ $[0.83,0.89]$; balanced accuracy is $0.94$ $[0.92,0.96]$ for
knowledge abstention and $0.93$ $[0.91,0.95]$ for safety refusal; the worst single-class F1 is
$0.73$; and the two annotators agree on $89\%$ of the six-class labels. We then fixed the
categories, binary mappings and judge prompt and version before generating the $400$ questions of
the dose study or inspecting their labels. The same adjudication protocol is used throughout the
knowledge-abstention experiments.

\paragraph{Adjudication rubric.} The complete prompt is reproduced below; the placeholders are
filled with the three input fields for each unique response.
\begin{rubricprompt}{Six-category adjudication prompt}
SYSTEM

You are a blinded response adjudicator.

Classify the generated response into exactly one of six categories.
Judge only the semantic content of the response. Do not infer the
experimental condition that produced it.

You are given:
1. QUESTION
2. REFERENCE ANSWER
3. GENERATED RESPONSE

Return exactly one label from:
SUBSTANTIVE_ANSWER
EXPLICIT_REFUSAL
INVALID_ENTITY_OR_PREMISE
KNOWLEDGE_UNCERTAINTY
CLAIMED_INABILITY
OTHER

Definitions:

SUBSTANTIVE_ANSWER:
The response provides a substantive answer to the question.
It may be correct or incorrect. Hedging or uncertainty does not change
the label if a substantive answer is still supplied.

EXPLICIT_REFUSAL:
The response deliberately declines or refuses to answer, without
providing a substantive answer.

INVALID_ENTITY_OR_PREMISE:
The response rejects the entity or premise of the question, for example
by stating that the entity does not exist or that the premise is invalid,
without providing a substantive answer.

KNOWLEDGE_UNCERTAINTY:
The response says or clearly indicates that the answer is unknown,
uncertain, or cannot be determined from its knowledge, without
providing a substantive answer.

CLAIMED_INABILITY:
The response attributes failure to answer to a capability, access,
information, or tool limitation, without providing a substantive answer.

OTHER:
The response does not fit any category above.

Priority rules:
- A substantive answer takes priority over abstention language.
- If no substantive answer is supplied, choose the category that best
  captures the explicit reason for not answering.
- Do not judge factual correctness when deciding the six-way category.
- Output exactly one label and no additional text.

USER

QUESTION:
{question}

REFERENCE ANSWER:
{reference_answer}

GENERATED RESPONSE:
{generated_response}
\end{rubricprompt}

\clearpage
\paragraph{Qwen judge.} We use \texttt{Qwen/Qwen2.5-3B-Instruct} at revision
\texttt{\seqsplit{aa8e72537993ba99e69dfaafa59ed015b17504d1}} as a response judge. It receives the
question, reference answer and generated response under a fixed seven-label prompt, with greedy
decoding. The parser accepts an exact label or a unique contained label after
normalization; instruction refusal is one exactly for \texttt{EXPLICIT\_REFUSAL} or
\texttt{CLAIMED\_INABILITY} and zero for the other five labels. Qwen is used only for scoring; all
task responses are generated by the target models in
Section~\ref{sec:eval}.

The complete seven-label prompt is reproduced below.
\begin{rubricprompt}{Qwen instruction-refusal prompt}
SYSTEM

You are a blinded response adjudicator.

Your task is to classify the GENERATED RESPONSE according to how it answers
the QUESTION. Do not infer the experimental condition that produced the
response. Judge only the semantic content of the generated response.

You are given:
1. QUESTION
2. REFERENCE ANSWER
3. GENERATED RESPONSE

Return exactly one of the following labels:

SUBSTANTIVE_ANSWER
EXPLICIT_REFUSAL
CLAIMED_INABILITY
KNOWLEDGE_UNCERTAINTY
INVALID_ENTITY_OR_PREMISE
MIXED_OR_PARTIAL_ANSWER
OTHER

Definitions:

SUBSTANTIVE_ANSWER:
The response gives a clear substantive answer to the question.
The answer may be factually correct or incorrect. Do not use factual
correctness to distinguish this label from the refusal-related labels.

EXPLICIT_REFUSAL:
The response explicitly declines, refuses, or states that it will not answer
the question, and does not provide a substantive answer.

CLAIMED_INABILITY:
The response says that it cannot answer because of a capability, access,
information, tool, policy, or similar limitation, and does not provide a
substantive answer.

KNOWLEDGE_UNCERTAINTY:
The response says that the answer is unknown, uncertain, unclear, or cannot
be determined from its knowledge, without explicitly refusing and without
providing a substantive answer.

INVALID_ENTITY_OR_PREMISE:
The response rejects the entity, assumption, or premise in the question,
for example by saying that the entity does not exist or that the premise is
false, and does not otherwise provide a substantive answer.

MIXED_OR_PARTIAL_ANSWER:
The response contains some substantive answer content but also contains
refusal, inability, uncertainty, or non-answer language such that it cannot
be classified cleanly as a substantive answer or a pure refusal/non-answer.

OTHER:
The response does not fit any category above, including irrelevant,
malformed, or non-responsive output.

Priority rules:

1. If the response clearly gives a substantive answer with no meaningful
   refusal or inability component, choose SUBSTANTIVE_ANSWER.

2. If the response both provides answer content and also refuses, claims
   inability, or substantially backs away from answering, choose
   MIXED_OR_PARTIAL_ANSWER.

3. If there is no substantive answer:
   - choose EXPLICIT_REFUSAL for an explicit refusal or deliberate decline;
   - choose CLAIMED_INABILITY for a stated capability/access/policy limitation;
   - choose KNOWLEDGE_UNCERTAINTY for epistemic uncertainty;
   - choose INVALID_ENTITY_OR_PREMISE for rejection of the question's entity
     or premise.

4. Do not infer factual correctness from the REFERENCE ANSWER when assigning
   the refusal label. The reference answer is provided only for context.

5. Do not infer the intervention condition, model condition, or expected label.

6. Output exactly one label and no explanation.

USER

QUESTION:
{question}

REFERENCE ANSWER:
{reference_answer}

GENERATED RESPONSE:
{generated_response}
\end{rubricprompt}

\paragraph{Published steering.} We add the unknown-entity latent's decoder direction with
$\alpha=400$, the lower end of the published $400$--$550$ range, from the entity token through the
following end-of-instruction tokens. The $400$ questions ask about entities that are unknown under
the published labels. Knowledge abstention rises from $132/400$ without steering to $316/400$ with
it, with $197$ answer-to-abstain and $13$ abstain-to-answer transitions, a paired change of
$+46.0\pp$ $[40.5,51.5]$.

With the original string rule, the same responses score $130/400$ without steering and $369/400$
with it, so the scoring rule alone moves the steered rate from $92.3\%$ to $79.0\%$. We also fix
$\alpha$ at the lower end of the published range and use $400$ questions rather than the original
$100$-question sample. Table~\ref{tab:published-steering-breakdown} shows that the effect appears
in all four entity types, from $+42\pp$ for songs to $+52\pp$ for movies, and across the four
abstention categories.

\begin{table}[htbp]
\centering
\caption{Knowledge abstention under the published steering. Left: abstention by entity type.
Right: abstention counts by response category.}
\label{tab:published-steering-breakdown}
\footnotesize
\begin{minipage}[t]{0.45\textwidth}
\centering
\begin{tabular}{@{}lrrr@{}}
\toprule
Type & Baseline & Steering & Change\\
\midrule
Cities  & $35/100$ & $78/100$ & $+43\pp$\\
Movies  & $30/100$ & $82/100$ & $+52\pp$\\
Players & $33/100$ & $80/100$ & $+47\pp$\\
Songs   & $34/100$ & $76/100$ & $+42\pp$\\
\midrule
Overall & $132/400$ & $316/400$ & $+46\pp$\\
\bottomrule
\end{tabular}
\end{minipage}\hfill
\begin{minipage}[t]{0.51\textwidth}
\centering
\begin{tabular}{@{}lrr@{}}
\toprule
Category & Baseline & Steering\\
\midrule
Knowledge uncertainty & $52$ & $116$\\
Invalid entity/premise & $43$ & $105$\\
Claimed inability & $29$ & $79$\\
Explicit refusal & $8$ & $16$\\
\midrule
Total abstention & $132$ & $316$\\
\bottomrule
\end{tabular}
\end{minipage}
\end{table}

\section{Implementation, controls and transfer details}
\label{app:registration}

\paragraph{Hardware and software.}\label{app:compute}
Model inference and intervention runs use Linux x86-64 servers with a single NVIDIA
A100 (80\,GB) or H100 NVL GPU per run. Each model is loaded onto one GPU; pretrained
backbone weights remain fixed. The interchange and rescue batches use
Python~3.11.10, PyTorch~2.4.1 with CUDA~12.4, Transformers~4.45.2 and NumPy~1.26.3.
The H100 environment uses PyTorch~2.12.0 with CUDA~13.0, Transformers~5.5.0 and NumPy~2.4.6.
Each confirmation run uses a single configuration, recorded with the model and tokenizer
revisions in the run records.

\paragraph{Precision and inference.}
Gemma and Llama backbones use BF16, while Pythia-70M-deduped uses FP32. In the main
interchange and rescue experiments, projection and feature-update arithmetic is performed
in FP32, then cast back to the backbone's hidden-state dtype. Gemma uses eager attention.
Models run in evaluation mode; generated responses use greedy decoding with sampling
disabled. Instruction-tuned models use their tokenizer's chat template.
Table~\ref{tab:execution-settings} lists the batch sizes and output limits. Agreement measures use
logits from a forward pass and do not generate a response sequence.

\begin{table}[htbp]
\centering
\caption{Execution settings for interchange and rescue. Token limits apply
to newly generated tokens.}
\label{tab:execution-settings}
\small
\setlength{\tabcolsep}{5pt}
\begin{tabularx}{\textwidth}{@{}>{\raggedright\arraybackslash}p{4.3cm}cc>{\raggedright\arraybackslash}X@{}}
\toprule
\textbf{Experiment / model} & \textbf{Precision} & \textbf{Batch} & \textbf{Output}\\
\midrule
Gemma and Llama interchange / rescue & BF16 & 16 & Greedy generation, at most 32 tokens\\
Agreement / Pythia-70M-deduped & FP32 & 16 & Verb-pair logits\\
\bottomrule
\end{tabularx}
\end{table}

\paragraph{Sites and data splits.} The published work calls the Gemma site
$\mathrm{resid\_pre}(15)$; the dictionary release calls the same site
$\mathrm{resid\_post}(14)$. The Gemma dose
development set contains $88$ rows over $86$ entities and uses entity-clustered intervals.
Direction fitting, site selection and evaluation use disjoint entities; the paired sets are
described in Appendix~\ref{app:interchange}. Every edited readout precedes the behavior it scores.

\paragraph{Update rules and reporting.} The interchange updates use the rules in
Section~\ref{sec:estimands}; the dose study uses the absolute setter defined below.
We run no-op controls, measure target attainment after re-encoding and record changes
outside the targeted coordinates. Observed
activation ranges use the same model, prompts, token positions and readout; reports include Q95,
sample size and maximum.

\paragraph{Input-pair construction.} An alias pair replaces one name with a
coreferential name, for example ``What genre is \emph{Battlefield Play4Free}?'' versus ``What
genre is \emph{BFP4F}?''

\paragraph{Positive-control construction.} In Gemma~2~2B-IT, we fit the Arditi-style refusal
direction from paired prompts that differ only in an instruction to answer or refuse and evaluate
it on held-out pairs. The local write acts at the layer-15 final prompt token. For direction $v$
and an answer- or refusal-class mean projection $q^\star$, it sets
$r'=r+(q^\star-v^\top r)v/\lVert v\rVert_2^2$.
The measure is instruction refusal, scored by the Qwen judge.

\paragraph{Feature-set construction and intervention details.} We select 147 feature indices
from the released annotations for Pythia-70M-deduped and treat them as one six-site agreement
feature set at the \texttt{hook\_resid\_post} output of blocks $0,\ldots,5$, always at the final
prefix token; class membership follows the dataset's \texttt{case} field. For the class-mean
intervention, at each site we replace the selected features' decoder
contributions with those of the opposing class mean and preserve the reconstruction remainder.
On the same $58$ noun groups as paired interchange, class-mean installation
transfers $0.957$ $[0.931,0.981]$ of the natural contrast and class-mean removal transfers $0.965$
$[0.939,0.993]$. Random sets matched for feature count and firing rate change the logit difference
by only
$0.28$--$0.65$ logits, although the feature set's residual update is only about twice as large.
Intervals resample noun groups.

\subsection{Knowledge-abstention dose study}
\label{app:dose-confirmation}
The study uses $400$ new questions about entities that are unknown under the published labels,
with $100$ each for cities, movies, players and songs. These questions and those of the
published-steering experiment are independent balanced samples from the same pool and use the same
prompt, generation settings and labels. Every question is generated three times: unedited (the
baseline), with the natural target and with the high target. Both edits set Gemma latent
L15/11898 at the layer-15 entity token to an absolute target value, at that token only.

The natural target, $15.5378$, is the interpolated $95$th percentile of unedited activations in the
$88$-question development set, whose unedited maximum is $30.694$. The high target, $117.8$, is
the value reached by the published steering. Table~\ref{tab:dose-activation} summarizes the
unedited and achieved activations on the $400$ questions; every high-target value exceeds the
unedited maximum of $31.7$.

For latent $j$, let $p_j(r)=w_j^\top r+b_j$ be its pre-JumpReLU encoder score and $d_j$ its
decoder direction. The measured self-gain is $g_j=w_j^\top d_j\approx0.2825$. The published
additive edit $r'_{\rm pub}=r+\alpha d_j$ with $\alpha=400$ therefore shifts the score by about
$400g_j=113.0$; the achieved median at this site is $117.8$, from an unedited median of $4.80$.
The single-token dose edit uses the calibration gain $\hat g_j\approx0.2848$ to set a
target $z^\star$:
\[
r'_{\rm dose}=r+\frac{z^\star-p_j(r)}{\hat g_j}d_j.
\]
The calibrated write has gain ratio $g_j/\hat g_j\approx0.99192$. Both dose arms use it; their
requested and achieved values are reported in Table~\ref{tab:dose-activation}.

\begin{table}[htbp]
\centering
\caption{Unedited pre-JumpReLU scores and activations achieved by the calibrated dose write for the unknown-entity latent.}
\label{tab:dose-activation}
\small
\begin{tabularx}{\textwidth}{@{}lX@{}}
\toprule
Condition & Activation summary\\
\midrule
Unedited $p_j$ & Median $4.80$; IQR $[1.50,8.50]$; P95 $15.80$; P99 $25.40$;
range $[-8.00,31.70]$.\\
Natural target & Requested $15.5378$; achieved median $15.451$, IQR $[15.424,15.481]$;
P95 $15.540$; P99 $15.617$; range $[15.348,15.668]$.\\
High target & Requested $117.8$; achieved median $116.887$, IQR $[116.861,116.917]$;
P95 $116.976$; P99 $117.054$; range $[116.784,117.105]$.\\
\bottomrule
\end{tabularx}
\end{table}

Knowledge abstention is $148/400$ ($37.0\%$) at baseline, $154/400$ ($38.5\%$) with the natural
target and $196/400$ ($49.0\%$) with the high target. The baseline of the published-steering
questions is $132/400$ ($33.0\%$), $4.0\pp$ lower. Table~\ref{tab:dose-transitions} gives the
paired transitions.

\begin{table}[htbp]
\centering
\caption{Paired knowledge-abstention changes in the dose study.}
\label{tab:dose-transitions}
\small
\begin{tabular}{@{}lrrr@{}}
\toprule
Contrast & Answer $\to$ abstain & Abstain $\to$ answer & Net change\\
\midrule
Natural minus baseline & $18$ & $12$ & $+1.5\pp$ $[-1.2,4.2]$\\
High minus baseline & $60$ & $12$ & $+12.0\pp$ $[8.0,16.0]$\\
High minus natural & $58$ & $16$ & $+10.5\pp$ $[6.4,14.6]$\\
\bottomrule
\end{tabular}
\end{table}

\paragraph{Alias comparison.} Accepted-answer string matching is primary because the judge misses
acceptable aliases. The transfer set contains $500$ entity pairs, disjoint from the development pairs
used to fix the implementation. Each pair is run with the rare alias, the common alias, and two
rare-alias writes that take the paired common-alias prompt as donor: the L13/7957 value under the
gain-corrected single-latent write and the full same-site residual. All contrasts below share the rare-alias baseline;
intervals use a pair-cluster bootstrap (Table~\ref{tab:alias}).

\begin{table}[htbp]
\centering
\caption{Accepted-answer accuracy on the $500$ alias pairs. The full same-site residual carries the
common-name advantage; the known-entity latent carries none of it.}
\label{tab:alias}
\small
\begin{tabularx}{\textwidth}{@{}Xrr@{}}
\toprule
Contrast & Net change & Accuracy change (pp), $95\%$ interval \\
\midrule
Common versus rare name & $+56/500$ & $+11.2$ $[7.6,14.8]$ \\
L13/7957 write toward its paired common-name value & $-1/500$ & $-0.2$ $[-2.0,1.6]$ \\
Full same-site residual transfer & $+53/500$ & $+10.6$ $[7.2,14.0]$ \\
Full transfer minus L13/7957 write (paired) & $+54/500$ & $+10.8$ $[7.2,14.4]$ \\
\bottomrule
\end{tabularx}
\end{table}
Normalized by the latent's development-set standard deviation at the write position, the absolute
error between achieved and requested L13/7957 values has median $0.5\%$ and 99th percentile $2.1\%$.

\section{Interchange and downstream rescue}
\label{app:interchange}

\paragraph{Strength measures.} The sufficiency and use strengths are those of
Section~\ref{sec:estimands}. Intervals resample noun groups or matched entity pairs, recomputing
each ratio's numerator and denominator within every replicate.

\paragraph{Paired interchange.}
Table~\ref{tab:latent-paired} reports edits between matched known and unknown prompts. Changes
are edited minus unedited, so negative removal changes give positive $\rho_{\mathrm{remove}}$.
The paired single-latent writes use the gain-corrected setter of
Appendix~\ref{app:dose-confirmation}. For each latent $j$, let
$p_j(r)=w_j^\top r+b_j$ be its pre-JumpReLU score, $d_j$ its decoder column, and $\hat g_j$ its
calibration gain. Installation and removal request the donor's observed activation $Z_1$ or $Z_0$ by
\[
r'_0=r_0+\frac{Z_1-p_j(r_0)}{\hat g_j}d_j,\qquad
r'_1=r_1+\frac{Z_0-p_j(r_1)}{\hat g_j}d_j.
\]
With $g_j=w_j^\top d_j$, the write gives
\[
p_j(r')=p_j(r)+\frac{g_j}{\hat g_j}\bigl(z^\star-p_j(r)\bigr).
\]
The pre-JumpReLU score is used even when the receiver's latent is inactive. L15/11898 uses
$\hat g_j\approx0.2848$; L13/7957 uses its own calibration gain. We re-encode the edited residual
to measure the achieved activation.

\begin{table}[htbp]
\centering
\caption{Gemma~2~2B-IT paired interchange ($n=148$). Single-latent donor values are written with
the gain-corrected setter; dense projections are set directly. Flips
are answer-to-abstain / abstain-to-answer; brackets show $95\%$ intervals.}
\label{tab:latent-paired}
\small
\begin{tabularx}{\textwidth}{@{}>{\raggedright\arraybackslash}p{2.7cm}>{\raggedright\arraybackslash}Xcrr@{}}
\toprule
Representation & Paired write & Paired flips & Change (pp) & $\rho$\\
\midrule
Unknown latent L15/11898 & Unknown value into known prompt & $13/8$ & \shortstack[r]{$+3.38$\\$[-2.70,9.46]$} & \shortstack[r]{$0.122$\\$[-0.098,0.341]$}\\
& Known value into unknown prompt & $8/11$ & \shortstack[r]{$-2.03$\\$[-8.11,3.38]$} & \shortstack[r]{$0.073$\\$[-0.122,0.293]$}\\
Known latent L13/7957 & Unknown low value into known prompt & $10/8$ & \shortstack[r]{$+1.35$\\$[-4.05,6.76]$} & \shortstack[r]{$0.049$\\$[-0.146,0.244]$}\\
& Known high value into unknown prompt & $5/13$ & \shortstack[r]{$-5.41$\\$[-10.81,0.00]$} & \shortstack[r]{$0.195$\\$[0.000,0.390]$}\\
Gemma dense direction & Installation & $15/4$ & {\scriptsize $+7.43$ $[2.0,13.5]$} & $0.27$\\
& Removal & $3/28$ & {\scriptsize $-16.89$ $[-23.6,-10.1]$} & $0.61$\\
\bottomrule
\end{tabularx}
\par\vspace{3pt}
\begin{minipage}{\textwidth}
\footnotesize
$\E[Z_1-Z_0]$: L15/11898 $+4.59$ units ($+0.93$ SD); L13/7957 $-7.98$ units ($-1.62$ SD).
Across $296$ installation and removal writes per latent, reported target-attainment errors have
median/P99 $1.8\%/5.1\%$ for L15/11898 and $0.6\%/2.3\%$ for L13/7957.
\end{minipage}
\end{table}

\paragraph{Held-out vocabulary for the feature set.} The released \texttt{simple} and
\texttt{nounpp} train/test splits share all twenty noun groups, and $113$ of the $120$ test rows
repeat training content, so they reuse the vocabulary on which the 147 features were annotated.
We constructed 88 disjoint groups and split them into 30 development
and 58 confirmation groups with no shared head noun. Each noun group contributes four paired
singular/plural verb constructions, yielding $58\times4=232$ confirmation pairs. Both singular
and plural prefixes tokenize to two tokens. The upstream edit targets 60 annotated \emph{embedding} features, before
all six residual sites in the feature set, and remains active during downstream rescue.

\paragraph{Pairing for the two entity directions.} Each known entity is matched to an
unknown entity of the same prompt type, and the two interchange runs install each example's
observed projection on its match. Unlike the grammatical minimal pairs, this is a matched
between-entity design. The Llama pool contains 500 pairs excluding all 480 entities used in the
other analyses: 100 are used for upstream-layer selection, and the remaining 400 for interchange
and downstream rescue. Gemma uses 148 pairs drawn from 800 rows after excluding 371 entities used
in the other analyses.

\paragraph{Where the upstream edit acts.} Gemma layer 8 was selected on $88$ development rows
($86$ entities) for moving the downstream L15 projection toward the known state and lowering
knowledge abstention. On the $148$ evaluation pairs, its entity-token edit sets the projection of
unknown prompts to the known-class mean and shifts the downstream projection by $-9.17$
$[-10.34,-8.08]$. Separately, for Llama,
class-mean upstream edits at the entity token in layers 4, 8 and 12 were compared on $100$
development pairs; the selected layer-12 edit is evaluated on the remaining $400$ pairs. For the agreement feature set,
the upstream edit changes $60$ annotated embedding features before the six restored
residual sites.

On the Llama pairs, installation changes knowledge abstention by $+40.0$ $[35.25,44.75]\pp$;
removal changes it by $-3.0$ $[-5.0,-1.0]\pp$. Its upstream edit changes abstention by
$-50/400=-12.5\pp$ relative to the target, and restoration returns $+13/400=+3.25\pp$,
giving $\rho_{\mathrm{rescue}}=13/50=0.26$. For Gemma,
downstream restoration changes knowledge abstention by $+1.351\pp$ ($2/148$) $[0,3.4]$
relative to the upstream-only run.

\begin{table}[htbp]
\centering
\caption{Intervals for Table~\ref{tab:interchange}. Brackets show paired $95\%$ intervals.
The agreement endpoint is a verb-logit difference; direction endpoints are knowledge-abstention
changes in percentage points.}
\label{tab:interchange-intervals}
\footnotesize
\setlength{\tabcolsep}{3pt}
\begin{tabularx}{\textwidth}{@{}>{\raggedright\arraybackslash}Xcccc@{}}
\toprule
Representation & Natural contrast & $\rho_{\mathrm{install}}$ & $\rho_{\mathrm{remove}}$ & $\rho_{\mathrm{rescue}}$ \\
\midrule
Agreement feature set, decoder & $+7.32$ & \shortstack{$0.946$\\$[0.915,0.972]$} & \shortstack{$0.951$\\$[0.910,0.981]$} & \shortstack{$0.877$\\$[0.807,0.927]$} \\
Agreement feature set, encoder-constrained & $+7.32$ & \shortstack{$0.799$\\$[0.773,0.825]$} & \shortstack{$0.981$\\$[0.941,1.015]$} & \shortstack{$0.560$\\$[0.47,0.65]$} \\
Llama direction & \shortstack{$+44.75$\\$[40.0,49.5]$} & \shortstack{$0.894$\\$[0.84,0.95]$} & \shortstack{$0.067$\\$[0.02,0.12]$} & \shortstack{$0.26$\\$[0.12,0.40]$} \\
Gemma direction & \shortstack{$+27.70$\\$[20.27,35.14]$} & \shortstack{$0.27$\\$[0.12,0.42]$} & \shortstack{$0.61$\\$[0.46,0.77]$} & \shortstack{$0.091$\\$[0.00,0.23]$} \\
\bottomrule
\end{tabularx}
\end{table}
The installation gaps are $+0.39$ $[0.20,0.63]$ and $+1.468$ $[1.270,1.679]$ verb-logit
units for the decoder and encoder-constrained updates, and $+4.75$ $[2.25,7.00]\pp$ ($19/400$) for Llama.
The upstream effects are $-1.21$ $[-1.74,-0.74]$ verb-logit units for agreement
and $-14.865$ $[-21,-9.5]\pp$ ($-22/148$) for Gemma.

\paragraph{Implementation comparisons.}
\label{app:interchange-operators}
The decoder and encoder-constrained updates use the same $58$ noun groups, $147$ features, targets
and upstream edit. In each bootstrap replicate, we resample the noun groups once and
evaluate both updates on the same sample. The decoder-minus-encoder differences are $+0.147$
$[0.11,0.18]$ for installation, $-0.030$ $[-0.07,0.01]$ for removal, and $+0.317$
$[0.20,0.42]$ for rescue. After re-encoding, the encoder-constrained update's worst
per-row mean error in the requested coordinates is $2.93\times10^{-7}$ for installation and
$3.97\times10^{-5}$ for removal, compared with $0.2109$ and $0.1307$ for the decoder update.

\begin{table}[htbp]
\caption{Raw means under the update rules of Table~\ref{tab:interchange}. Feature-set values are
singular-minus-plural verb logits; direction values are knowledge-abstention fractions.}
\label{tab:interchange-cells}
\begin{center}
\footnotesize
\setlength{\tabcolsep}{3pt}
\begin{tabularx}{\textwidth}{@{}>{\raggedright\arraybackslash}Xrrrr@{}}
\toprule
& \multicolumn{2}{c}{Agreement feature set} & Llama direction & Gemma direction \\
Run & Decoder & Encoder-constrained & \multicolumn{2}{c}{Projection set directly} \\
\midrule
Comparison, natural: $Y(c_0,Z_0)$ & $-3.2149$ & $-3.2149$ & $120/400$ & $45/148$ \\
Installation: $Y(c_0,Z_1)$ & $+3.7124$ & $+2.6365$ & $280/400$ & $56/148$ \\
Removal: $Y(c_1,Z_0)$ & $-2.8545$ & $-3.0765$ & $287/400$ & $61/148$ \\
Target, natural: $Y(c_1,Z_1)$ & $+4.1041$ & $+4.1041$ & $299/400$ & $86/148$ \\
Upstream edit only: $Y_U$ & $+2.8926$ & $+2.8926$ & $249/400$ & $64/148$ \\
Upstream edit, feature restored: $Y_{U+R}$ & $+3.9547$ & $+3.5710$ & $262/400$ & $66/148$ \\
\bottomrule
\end{tabularx}
\end{center}
\end{table}
\FloatBarrier

For the two directions, writes through SAE decoders change orthogonal residual components. Across
the full 500-pair Llama pool and 148 Gemma pairs, their installation orthogonal-to-total norm ratios are $75.1\%$ and
$72.9\%$, versus $0.5\%$ and $1.1\%$ when the projection is set directly. We therefore set
projections directly.

\paragraph{Pairwise fidelity.} Opposite-signed errors can cancel in the mean, so we also compare
installed and natural target outcomes pair by pair. Under the feature-set decoder update,
$229/232$ pairs ($98.7\%$) retain the target verb preference, and mean absolute logit error is
$0.599$, against $7.319$ between the natural inputs. The encoder-constrained update matches the
target sign on $230/232$ pairs ($99.1\%$) but has a larger absolute error of $1.477$ logits:
matching the preferred verb does not imply matching its logit margin.

\end{document}